\documentclass[twoside,leqno,twocolumn]{article}

\usepackage[letterpaper]{geometry}
\usepackage[table]{xcolor}
\usepackage{url}
\usepackage{graphicx,amsmath}
\usepackage{ltexpprt}
\usepackage{multirow}
\begin{document}

\title{\Large BKT-LSTM: Efficient Student Modeling \\ for knowledge tracing and student performance prediction}
\author{Sein Minn\thanks{Inria - Lille Nord Europe, France}}

\date{}

\maketitle

\begin{abstract} \small\baselineskip=9pt

Recently, we have seen a rapid rise in usage of online educational platforms. The personalized education became crucially important in future learning environments. Knowledge tracing (KT) refers to the detection of students' knowledge states and predict future performance given their past outcomes for providing adaptive solution to Intelligent Tutoring Systems (ITS). Bayesian Knowledge Tracing (BKT) is a model to capture mastery level of each skill with psychologically meaningful parameters and widely used in successful tutoring systems. However, it is unable to detect learning transfer across skills because each skill model is learned independently and shows lower efficiency in student performance prediction. While recent KT models based on deep neural networks shows impressive predictive power but it came with a price. Ten of thousands of parameters in neural networks are unable to provide psychologically meaningful interpretation that reflect to cognitive theory. In this paper, we proposed an efficient student model called BKT-LSTM. It contains three meaningful components: individual \textit{skill mastery} assessed by BKT,  \textit{ability profile} (learning transfer across skills) detected by k-means clustering and \textit{problem difficulty}. All these components are taken into account in student's future performance prediction by leveraging predictive power of LSTM. BKT-LSTM outperforms state-of-the-art student models in student's performance prediction by considering these meaningful features instead of using binary values of student's past interaction in DKT. We also conduct ablation studies on each of BKT-LSTM model components to examine their value and each component shows significant contribution in student's performance prediction. Thus, it has potential for providing adaptive and personalized instruction in real-world educational systems.

\end{abstract}

\section{Introduction}

Knowledge Tracing (KT) is the assessment of students' knowledge state dynamically and the prediction of whether students may or may not answer a problem correctly based on past item test outcomes. KT leverages machine learning and data mining techniques to provide better assessment, enabling supportive learning feedback and adaptive instructions.  By tracing the knowledge state of the students, we can optimize instruction for individualization and personalization. 

\textit{Knowledge Assessment} methods (student models) were initially proposed by psychometricians to analyse and predict student performance in static data (e.g. exam data) according to cognitive theory~\cite{templin2010diagnostic}. Early state-of-the-art methods are either based on modeling of temporal latent variables or factor analysis with temporal features. In recent years,  there has been increasing interest by researchers from computer science background to provide a solution to assess student's knowledge proficiency and predict student performance by using data mining, machine learning and deep learning techniques on dynamic data. A wide array of Artificial Intelligence (AI) and Knowledge Representation techniques have been utilized in the field of ITS~\cite{nkambou2010advances}. Most of the main AI techniques have found their way into the building of adaptive learning environments, and in particular for the problem of knowledge tracing, which aims to model the student's state of mastery of conceptual or procedural knowledge from observed performance on tasks~\cite{corbett1994knowledge}. Besides, it shows great success in student performance prediction than early student models~\cite{piech2015deep,minn2019dynamic}. Recurrent Neural Networks (RNNs) models, used in the Deep Knowledge Tracing (DKT) approach~\cite{piech2015deep}, are a more recent example. 

Bayesian Knowledge Tracing (BKT) is the earliest sequential approach to student modeling with psychologically meaningful parameters, which works on independent skills without considering learning transfer across skill. While Piech~\cite{piech2015deep} argued that DKT has the ability to model some of the learning transfer occurring across skills by taking all past student's interaction (binary values with associated skills). However all these information is in hidden layer of LSTM with ten of thousands of parameters and unable to provide psychologically meaningful interpretation~\cite{khajah2016deep}.DKT shows better predictive performance than the preceding approaches, in part because they are known to preserve past information in sequential data, such as student outcome traces. Thus, we proposed an efficient student model called BKT-LSTM, which contains three meaningful components :   \textit{skill mastery} of individual skills assessed by BKT,  \textit{learning transfer} detected by k-means clustering and  \textit{problem difficulty} and then takes these three components into account in student performance prediction by using LSTM.

This paper is organized as follows: we review knowledge assessment and knowledge tracing models in section~\ref{sec:back}. Then we propose an efficient knowledge tracing framework in section~\ref{sec:model}. We compare with state-of-the-arts knowledge tracing models and perform ablation studies in section~\ref{sec:experimental-study}. Finally, we conclude in section~\ref{sec:conclusion}. 

\section{Background}\label{sec:back}

In successful learning environments like  Cognitive tutors and the ASSISTments,  KT plays as a mechanism for tracing learners knowledge \cite{desmarais2012review}.  In these systems, each problem is labeled with underlying skills required to correctly answer that problem. KT can be seen as the task of supervised sequential learning problem. The KT model is given student past interactions with system that includes: skills $S=\{s_1,s_2,..,s_t\}$ \cite{minn2016refinement} along with responses $R=\{r_1,r_2,..,r_t\}$  and predict the probability of getting correct answer for next problem is mainly depends on mastery of corresponding skill $s$ associated with problems  $P=\{p_1,p_2,..,p_t\}$ . So we can define the probability of getting correct answer as $p(r_{t}=1|s_{t},X)$ where$X=\{x_1,x_2,..,x_{t-1}\}$ and $x_{t-1}= (s_{t-1},r_{t-1})$ is a tuple containing response $r$ to skill $s$ at time $t-1$. Then, we review here best known state-of-the-art KT modeling methods for estimating student's performance.

\subsection{Item Response Theory}\label{sec:IRT}

In standardized tests, students' proficiency is assessed by one static latent variable for example Rash Model ~\cite{hambleton1991fundamentals} which shows a strong theoretical background both in terms of being grounded in psychometric measurement and a sound mathematical framework. However, it is not possible to perform cognitive diagnosis. IRT takes dichotomous item response outcomes and assigns student $i$ with a proficiency $\theta_i$, which is not changing during the exam. Each item $j$ has its own difficulty $\beta_j$ .

The main idea of IRT is estimating a probability that student $i$ answers item $j$ correctly by using student ability and item difficulty,

\begin{equation} \label{equ:irt}
  P_{ij}=logistic(\theta_{i}-\beta_{j})\\
  = \frac{1}{1+e^{\theta_i - \beta_j}}
\end{equation}

Where $\theta$~is the student proficiency on the topic tested, and $\beta_j$ is the difficulty of item~$j$.

Recently Wilson et al. proposed a Bayesian extension of IRT (We will refer this model as BIRT) \cite{wilson2016back} which uses Bayesian approach and regularizes on $log P_{ij}$ by imposing independent standard normal prior distributions over each
$\theta_i$ and $\beta_j$.It maximizes on log posterior probability of $\{\theta_{i},\beta_{j}\}$ given the response data $\{r(i,j,t)\in D\}$, where response $r\in\{0,1\}$ and $t$ is the time of each attempt. 

\begin{equation} \label{equ:irt1}
\begin{split}
\log P(\{\theta_i\},\{\beta_j\}|D) &= \sum_{(i,j,r,t)\in D} r \log f(\theta_i -\beta_j) + (1-r)  \\ & \log (1-f(\theta_i - \beta_j))
- \frac{1}{2} \sum_i \theta_i^2 \\ & -\frac{1}{2}  \sum_j \beta_j^2  + C.
\end{split}
\end{equation}
BIRT leverages the information of both items and students that directly interact in the system. Maximum a posteriori (MAP) estimates of $\theta_{i}$ and $\beta_{j}$ are computed by Newton-Raphson method. BIRT shows competitive performance to Deep Knowledge Tracing (DKT, a neural networks based knowledge tracing model mentioned in section \ref{sec:DKT}) in term of student performance prediction.

\subsection{Bayesian Knowledge Tracing} \label{sec:BKT}

Bayesian Knowledge Tracing (BKT) is the earliest sequential approach to model a learner’s changing knowledge state and is arguably the first model to relax the assumption on static knowledge
states. In original BKT, a single skill is tested per item and the learner’s
state of skill mastery is inferred at each time step based on a sequence of her previous outcomes  \cite{corbett1994knowledge}. 
\begin{figure}[h]
	\centering \includegraphics[width=0.66\linewidth]{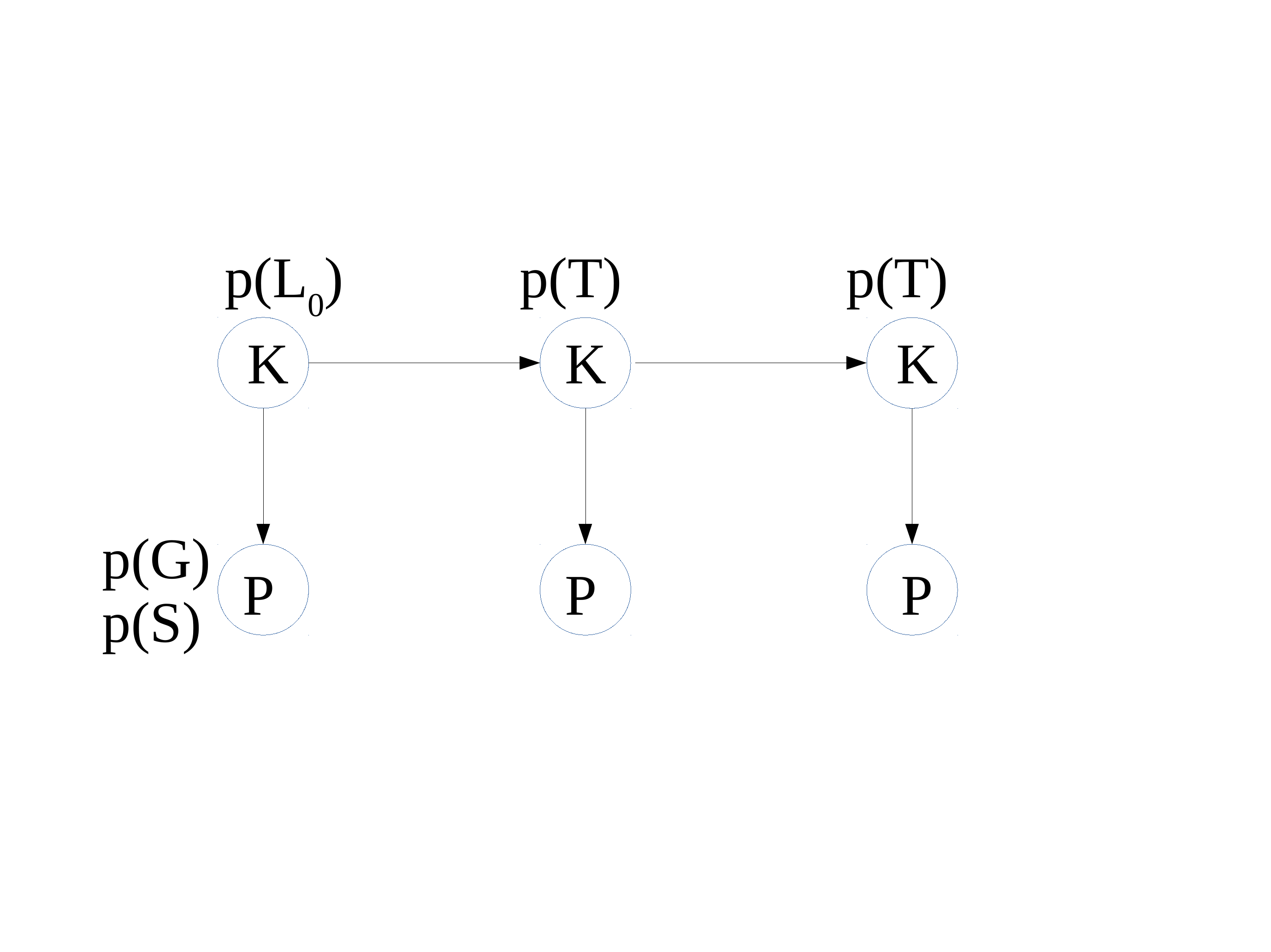} \caption{Architecture of BKT.}
	\label{fig:BKT} 
\end{figure}

BKT takes four probabilities:  $P(L_{0})$: the prior probability of a student knows the skill; $P(T)$: the probability a student, who does not currently know the skill, will know it after the next practice opportunity; $P(G)$: the probability a student guesses a question and gets a correct answer despite not knowing the skill (\textit{guess}); and  $P(S)$: the probability a student answers a question incorrectly despite knowing the skill (\textit{slip}).

In a typical learning environment with BKT, the estimate of student mastery of a skill is continually updated every time student gives a response  to an item. The probability that a student knows the skill at time $t$  is:

\begin{equation} \label{equ:corr}
P(L_{t}|1) = \frac{P(L_{t-1})(1-P(S))}{P(L_{t-1})(1-P(S))+(1-P(L_{t-1}))P(G)}
\end{equation}

\begin{equation} \label{equ:incorr}
P(L_{t}|0) = \frac{P(L_{t-1}) P(S)}{P(L_{t-1})P(S)+(1-P(L_{t-1}))(1-P(G))}
\end{equation}

\begin{equation} \label{equ:action}
P(L_{t}) = P(L_{t}|Action)+ (1-P(L_{t}|Action))P(T)
\end{equation}
And the probability of student applying the skill correctly with problem at time $t$ is computed as. 
\begin{equation} \label{equ:action}
P(C_{t}) = P(L_{t-1}) . (1-P(S))+ (1-P(L_{t-1})) . P(G)
\end{equation}

where 0 represents incorrect attempt and 1 represents correct attempt for $Action$ by student. First, the probability that a student knows the skill before the response is updated with $Action$ by using the equations \ref{equ:corr} or \ref{equ:incorr} according to the response evidence. Then, the system estimates the possibility that the student learns the skill during the problem step by using equation \ref{equ:action}. 

Several extensions of BKT have been introduced, such as the contextualization of estimates of guessing and slipping parameters, estimates of the probability of transition from the use of help features, and estimates of the initial probability that the student knows the skill. Item difficulty has also been integrated in this model.

\subsection{Performance Factors Analysis}\label{sec:PFA}

Performance Factor Analysis (PFA) was adapted from Learning Factor Analysis (LFA, \cite{cen2006learning}) to allow the creation of a “model overlay” that traces predictions for individual students with individual skills so as to provide the adaptive instruction to automatically remediate current performance \cite{pavlik2009performance}. 

Pavlik et al. adapted the LFA model with sensitivity to the indicator of student learning performance. The PFA model allows conjunction by summing the contributions from all skills needed in a performance. PFA also relaxes the static knowledge assumption and models multiple skills simultaneously~\cite{pavlik2009performance}. Its basic structure is:

\begin{equation}
\log P_{ij}= \beta_{j}+ \sum_{k\in KCs}(\gamma_{k}s_{ik}+\rho_{k}f_{ik}),\label{equ:pfm}
\end{equation}
where $\beta_{j}:$  the bias for the item $j$; $\gamma_{k}:$ the bias for success attempt to skill $k$; $\rho_{k}:$   the bias for failure attempt to skill $k$; $s_{ik}:$ the number of successful attempts of student $i$ on skill $k$; $f_{ik}:$     the number of failure attempts of student $i$ on skill $k$.

It does not consider student proficiency $\theta$, because it assumes that $\theta$
cannot be estimated ahead of time in adaptive situations \cite{pavlik2009performance}.

\subsection{Deep Knowledge Tracing} \label{sec:DKT}

Piech~\cite{piech2015deep} introduced Deep Knowledge Tracing (DKT) at the NIPS 2015 conference. Akin to BKT, it uses data in which skills are tried and the performance outcome to predict future sequence attempts. It encodes skill and student response attempt in a one-hot feature input vector as input for each time $t$. The output layer provides the predicted probability that the student would answer that particular next problem correctly at time $t+1$.

DKT uses Long Short-Term Memory (LSTM) to represent the latent knowledge space of students along with the number of practices dynamically. This model compactly encodes the historical information from previous time steps and determines what and how much information to remember by using input gate, forget gate and output gate of LSTM.

\begin{figure}[h]
	\centering \includegraphics[width=.5\linewidth]{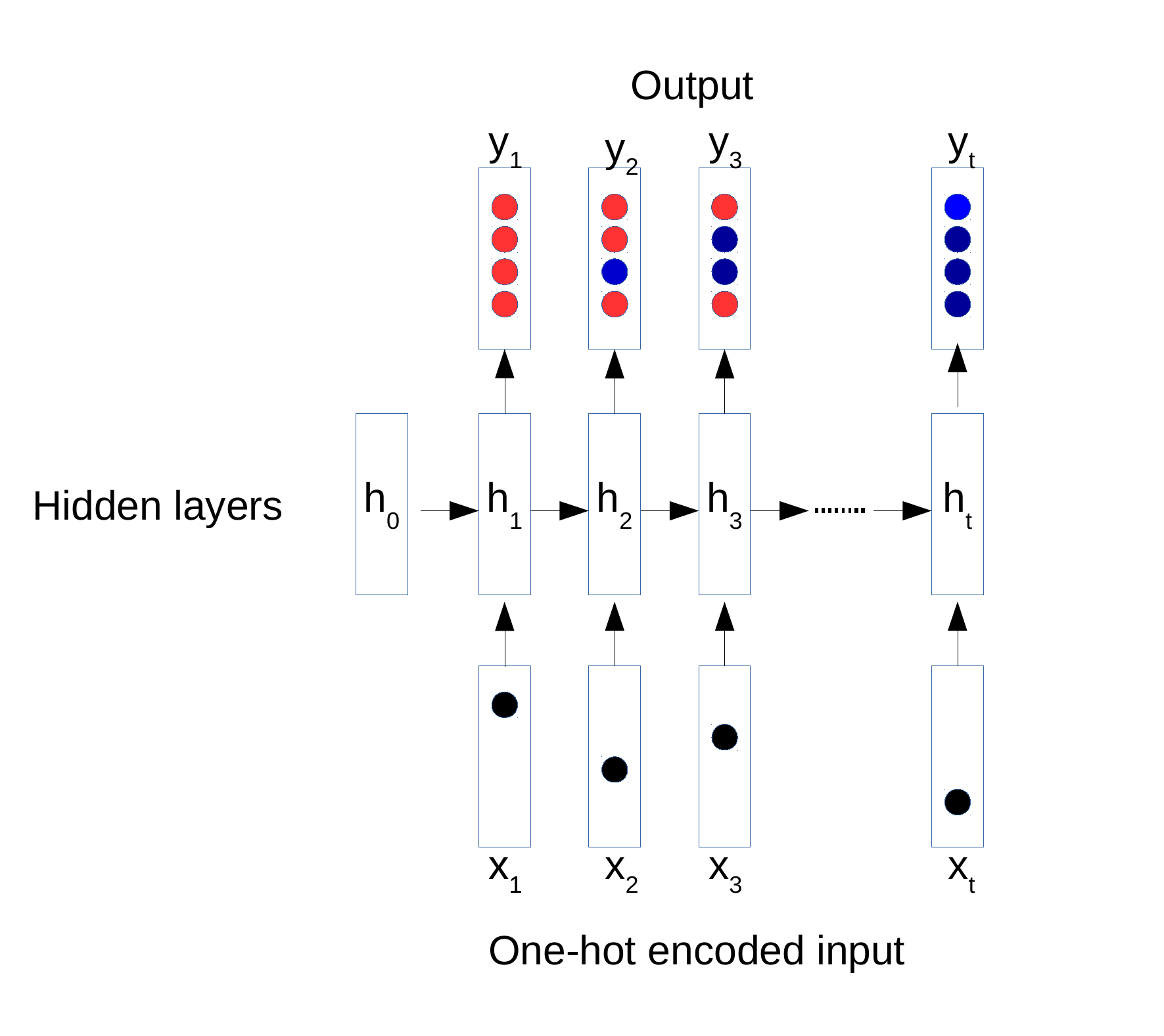} \caption{DKT unfolded architecture.}
	\label{fig:DKT} 
\end{figure}

DKT uses large numbers of artificial neurons for representing latent knowledge state along with a temporal
dynamic structure and allows a model to learn the student's  knowledge
state from data. It is defined by the following equations: 
\begin{equation}
h_{t}=\tanh(W_{hx}x_{t}+W_{hh}h_{t-1}+b_{h}),\label{equ:hidden}
\end{equation}
\begin{equation}
y_{t}=\sigma(W_{yh}h_{t}+b_{y}).\label{equ:output}
\end{equation}

In DKT, both tanh and the sigmoid function are applied element wise and parameterized by an input weight matrix $W_{hx}$, recurrent weight matrix $W_{hh}$, initial state $h_{10}$, and output weight matrix $W_{yh}$. Biases for latent and output units are represented by $b_{h}$ and $b_{y}$.

Where $x_t$ is a one-hot encoded vector of the student interaction $x_t = \{s_t, r_t\}$ that represents the combination of which skill $s_t$ was practiced with student response $r_t$, so $x_t \in \{0, 1\}^{2M}$ according to number $M$ of unique skills. The output $y_t$ is a vector of number of skills, where each value represents the predicted probability that the student would answer that particular problem with associated skill correctly. Thus the prediction of $r_{t+1}$ of next problem associated with skill $s_{t+1}$ can be retrieved from vector of $y_t$.

\subsection{Dynamic Key-Value Memory Networks for Knowledge Tracing} \label{sec:DKVMN}

\begin{figure}[h]
	\centering \includegraphics[width=.5\linewidth]{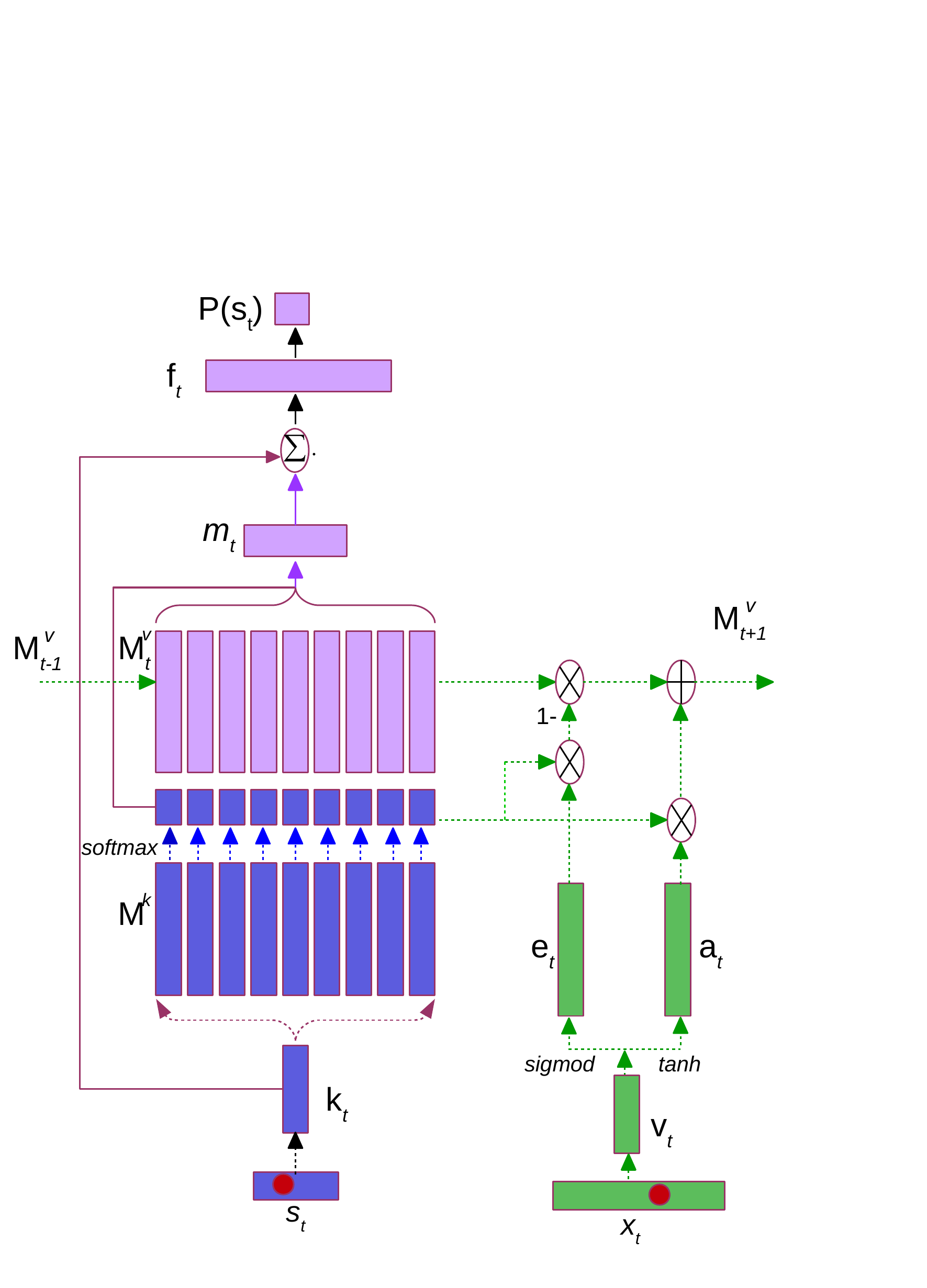} \caption{DKVMN architecture.}
	\label{fig:DKVMN} 
\end{figure}
DKVMN was proposed as an alternative to DKT that is inspired from the memory network architecture \cite{zhang2017dynamic}.  It utilises an external memory neural network module and uses two memory slots called key memory and value memory to encode the knowledge state of students. Assessments of knowledge state on particular skills are stored in memory slots and controlled by read and write operations through additional attention mechanisms. DKVMN calculates attention weight of inputs by using key memory, which is immutable, and reads and writes to the corresponding value memory. DKVMN takes a skill $s_t$ as query, estimate the probability of response $p(r_t|s_t)$ to the problem with associated skill $s_t$, and then updates the value memory with skill and response $x_t\{s_t, r_t\}$. Note that DKVMN consumes more memory storage than DKT.

The overall model architecture of DKVMN is shown in Figure \ref{fig:DKVMN}. Unlike DKT, DKVMN applies read and write operations to perform knowledge state transitions in memory rather than  unstructured state-to-state  knowledge transformation in hidden layer of DKT. Knowledge state of a student student is traced by reading and writing to the value memory slots using attention weight computed from input skill and the key memory slots. It comprises three main steps: Reading, Prediction and Writing. In addition to these three steps, two other processes should be mentioned, Optimization and Attention. See \cite{zhang2017dynamic} for more details.

In the training process, the initial value of both the key and the value
matrices are learned. Each slot of the key memory are embedded with concepts and is fixed in the testing process. 

\section{Proposed model: BKT-LSTM}\label{sec:model}

We propose a student model called BKT-LSTM which can predict student’s
future responses by utilizing three meaningful features: individual \textit{skill mastery} of a student assessed by BKT and \textit{learning transfer} (across skills) detected by k-means clustering and \textit{problem difficulty}. 

\subsection{Assessing skill mastery}\label{sec:skillmastery}

\begin{figure}[h]
	\centering \includegraphics[width=0.66\linewidth]{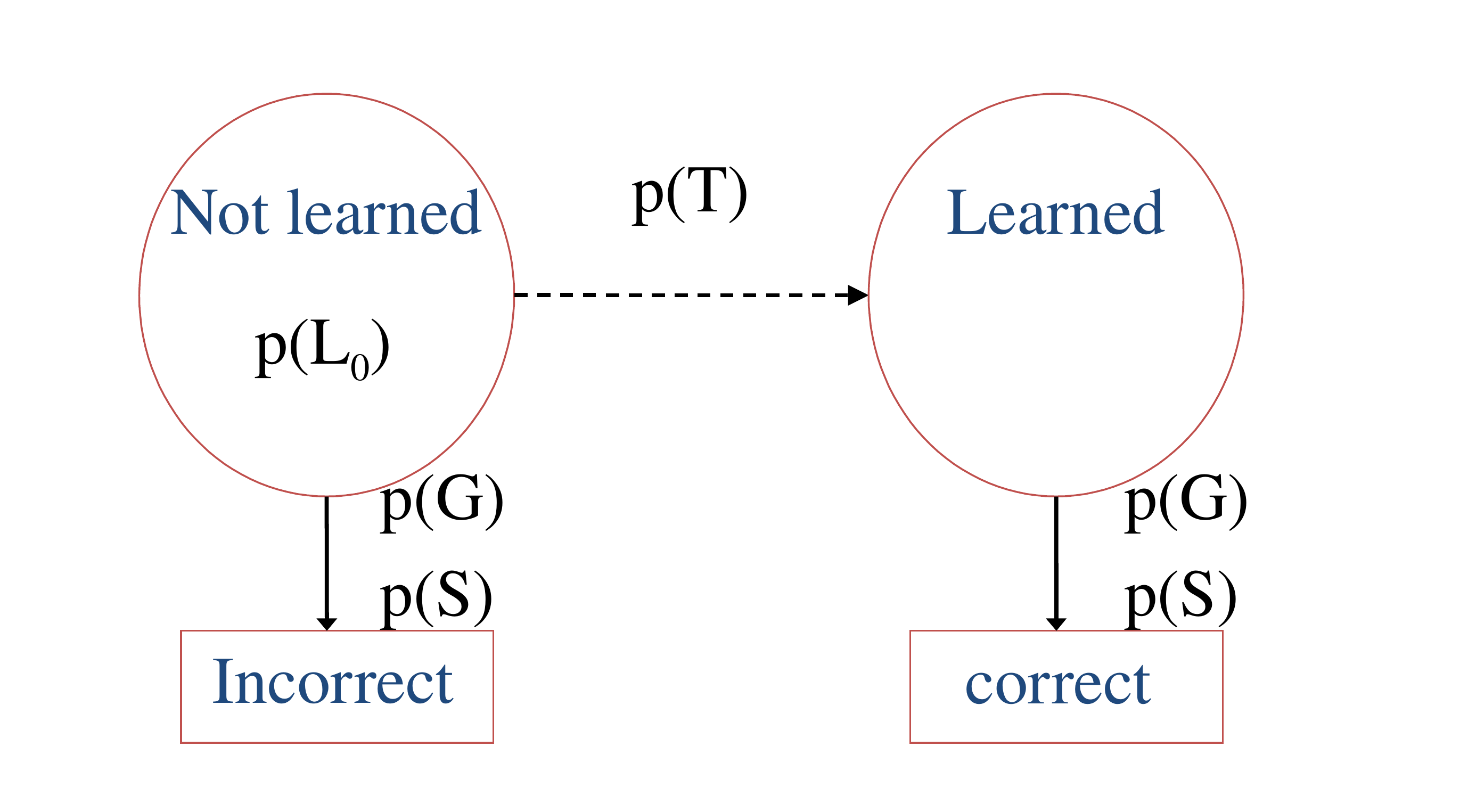} \caption{Example of standard BKT structure.}
	\label{fig:BKT1} 
\end{figure}

 The formulation is inspired by Bayesian Knowledge Tracing (BKT), a well known knowledge tracing model with psychologically meaningful parameters. BKT is a Markov model to infer mastery states, from “not learned" to “learned” and to the extent the probabilities above depend either on fixed parameters and on the state in the previous time step $t-1$ in Fig \ref{fig:BKT1}. As we mentioned in section \ref{sec:BKT}, we apply brute-force search algorithm to fit BKT. We have $P(L_{0}), P(T), P(G), P(S)$ of each skill after fitting BKT. BKT is based on skill specific modeling and able to provide skill mastery of each skill $P(s_t)= P(L_{t-1})$ according to:

\begin{equation} \label{equ:action}
P(L_{t-1})= P(L_{t-1}|Action)+ (1-P(L_{t-1}|Action))P(T)
\end{equation}

\textbf{Note that}: \textit{skill mastery} is not binary value and which is the probability of learning skill $s_t$ rather than the probability of student applying the skill correctly $P(C_{t})$ . Besides, each skill model is learned independently with four parameters. So, the time $t$ in BKT only represents number of trials student practised only on each specific skill. Other interleaved skills are ignored during modeling of that skill. So it doesn't consider learning transfer across skills.

\subsection{Learning Transfer}

A strong limitation of BKT is skill independent.  When a student is tried on a new skill, the learning she has gained on previous skills is not considered relevant to the new skill tried. This is known as the \textit{learning transfer} mechanism that is lacking in BKT and in most KT models.

Intuitively, learning transfer across skills is a naturally occurring phenomena. Most people would agree with the observation that general domain knowledge accumulates through practice and allows the student to solve new problems in the same domain with greater success, even if that problem involves new and specific skills. Learning transfer implies that students are able to transfer their acquired skills to new situations and across problems involving different skills set.  Models such as Learning Factor Analysis (LFA\cite{cen2006learning}) and Performance Factor Analysis (PFA\cite{pavlik2009performance}) aim to capture this learning transfer phenomena by including a factor that represents the learning accumulated on all KCs through practice and include this factor as a predictor of success in further practice.  These models have been shown to outperform the standard BKT model in general, and have given rise to further research on the subject. 

In BKT, learning transfer is ignored when the prediction is performed. So the model can not evaluate which type of learning transfer that student achieved at current time interval in long term learning process. Skill mastery is estimated independently without considering learning transfer across skills. In DKT, LSTM uses single state vector to encode the temporal information of student knowledge state with corresponding transfer learning and performs state-to-state transition globally and unable to assess the mastery of skills and it's ability profile for learning transfer.

\subsubsection{Detecting ability profile :}\label{sec:DTAB}

To detect the regularities on changes of learning transfer (ability profile of a student) over series of time intervals in long-term learning process,  we refined the method proposed in DKT-DSC \cite{minn2018deep} for better simplicity without scarifying it's originality and performance. It divides student's interactions into multiple time intervals, then encodes student past performance for estimating her ability profile in current time interval. The ability profile~$ab$ is encoded as a cluster ID and computed from the performance vector (with Equation~\ref{equ:seg}) of length equal to the number of skills, and updated after each time interval by using all previous attempts on each skill. The  success rates on each skill from past attempts data are transformed into a performance vector for clustering student $i$ at time interval $1\!\!:\!\!z$ as follows (for brevity we omit indexing all terms by $i$ in equation \ref{equ:segcorr}):

\begin{equation}
R(x_{j})_{1:z}=\sum_{t=1}^{z}\frac{(x_{jt})}{|N_{jt}|},\label{equ:segcorr}
\end{equation}
\begin{equation}
d_{1:z}^{i}=(R(x_{1})_{1:z},R(x_{2})_{1:z},...,R(x_{n})_{1:z}),\label{equ:seg}
\end{equation}

where 
\begin{itemize}
	\item $x_{jt}$: the attempts of skill~$x_{j}$ being correctly answered at time interval $t$; successful attempts are set to 1 and 0 otherwise;
	\item $|N_{jt}|$: the total number of practices of skill~$x_{j}$ up to time interval $z$;
	\item $n$: the total number of skills;
        \item $R(x_{j})_{1:z}$ represents ratios of skill~$x_{j}$ being correctly answered from time interval 1 to current time interval~$z$ by student~$i$. This is computed for all skills $(x_{1},x_{2},..,x_{n})$;
        \item $d_{1:z}^{i}$ represents a \textit{performance vector}  of student~$i$ on   all skills from time interval 1 until~$z$.
  
\end{itemize}

\begin{figure*}
	\centering \includegraphics[width=12cm]{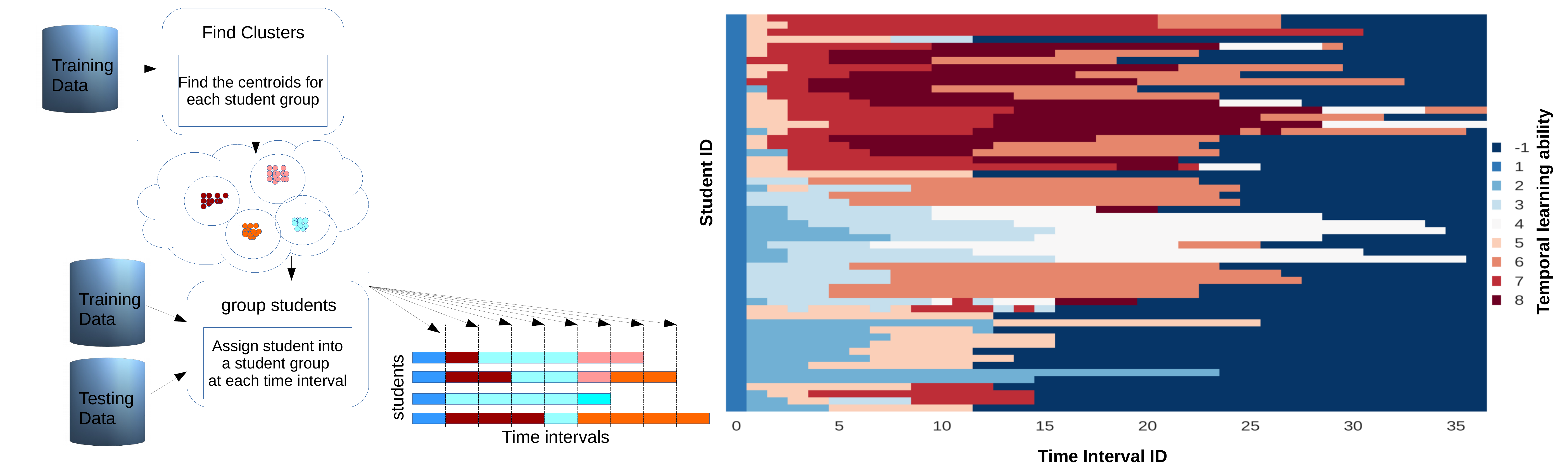} \caption{ Evaluation process of student's learning ability (Left) and Detecting ability profile at each time interval in long-term learning process of random 56 students in ASSISTments 2009 dataset (Right)}
	\label{fig:cluster} 
\end{figure*}

Each student may have a different number of total time intervals in the lifetime of their interactions with the system (see Figure.~\ref{fig:cluster}).

If a student's time interval has no attempt between interval $0:z$, we apply 0.5 for success rate, $R(x_{j})_{1:z}$.

So data contains encoded vector of student's past performance and it is accumulated and updated after each time interval. Time interval $z$  and student $i$ are ignored in training process and only used in clustering process later. Then k-means algorithm is used to evaluate the temporal long-term learning ability of students in both training and testing at each time interval $z$ by measuring euclidean distance with centroids achieved after training process as like in DKT-DSC \cite{minn2018deep} and assigning into nearest cluster and applied label of cluster $ab_z$ as temporal student learning ability at time interval $z$ Evalution is started after first 20 attempts and then after every 20 attempts have been made by a student. For first time interval, every student is assigned with initial ability profile 1 as described in Fig \ref{fig:cluster} (Right).

By adding this cluster ID $ab_{z}$ (ability profile) of what group the student belongs to, we ensure that these high-level skill profiles are available to the model for making its predictions throughout the long term interaction with the tutor

\subsection{Calculating problem difficulty}\label{sec:difficulty}

The integration of problem difficulty do predict student performance is a distinct feature. We calculate problem difficulty by using method suggested in \cite{minn2018improving,minn2019dynamic}. Calculation of the difficulty level is reconfigured for more details. Note that, in this study, we assume each problem is associated with a single skill, but the difficulty is associated with problems, not with skills themselves. 

The difficulty of a problem, $p_j$, is determined on a scale of 1 to 10.  This scale is used as an index into the one-hot vector: a value of 1 is set to the cell with the corresponding difficulty level, and 0 for all others.  Problem difficulty level $PD(P_j)$ is calculated as:
\begin{equation} \label{equ:incorrratio} 
\textbf{difficulty level}(p_j) = 
\begin{cases}
\delta(p_{j}), & \text{if}\ \rvert N_j \lvert \geq 4\\
5, & \text{else}
\end{cases}
\end{equation}
where:
\begin{equation} \label{equ:catdelta}
\delta(p_{j}) = \mathrm{modulo}_{10} \left( { \frac{{ \sum_i^{\lvert N_j \rvert }}{O_i(p_{j})}}{\lvert N_j \rvert}} \cdot 10 \right )
\end{equation}
and where 
\begin{itemize}
	\item $p_j$: problem~$j$
	\item $N_j$ is the set of students who attempted problem~$p_j$
	\item $O_i({p_{j}})$ the outcome of the first attempt from student~$i$ to problem~$p_j$, 1~if a success, 0~otherwise

\end{itemize}

$\delta(p_{j})$ is a function that maps the average success rate of problem~$p_j$ onto $(10)$ levels. Unseen problems, those that do not have any record, and problems with less than 4~students ($ \rvert N_j\lvert < 4$) in the dataset, will have a difficulty level of 5.

\begin{figure}
	\centering \includegraphics[width=.7\linewidth]{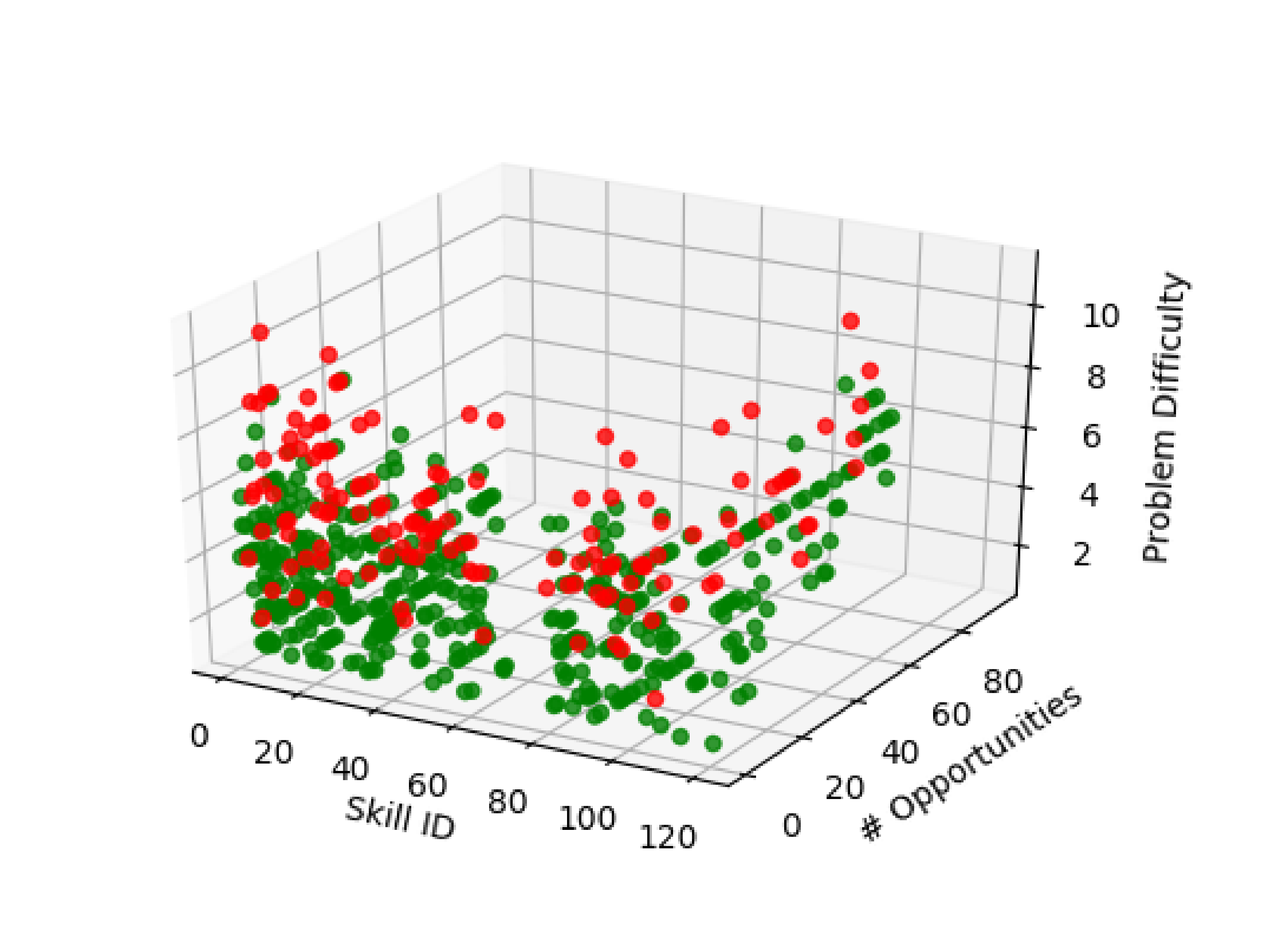} \caption{ a student's responses in different levels of problem difficulty at each trial: red is incorrect response and green is a correct response.}
	\label{fig:Stu_Diff}
\end{figure}

Figure \ref{fig:Stu_Diff} shows the distributions of successes and failures over skills as a function of problem difficulty, and along the number of opportunities of practice over that skill.  As expected, more failures are found at the higher difficulty levels, and also at early attempts over a given skill.  Notice also that failures after a number of attempts on a given skill do occur, but only at the higher problem difficulty levels.  These results demonstrate that problem difficulty does play an important role on the chances of success or failures to a problem, even for well practiced skills.

\subsection{BKT-LSTM} 
To handle all deficiencies described in above, we  proposed a novel model called BKT-LSTM by leveraging advantageous of predictive power of LSTM.  BKT-LSTM contains three meaningful components: skill mastery assessed by BKT, learning transfer evaluated by k-means clustering and problem difficulty. BKT-LSTM  predict student performance based on both of evaluated temporal  student's ability profile that capture \textit{learning transfer}, assessed \textit{mastery of skills} from BKT simultaneously and take  difficulty level of a problem in performance prediction. In BKT-LSTM, we defined LSTM by following equations: 
\begin{equation}
h_{t}=\tanh(W_{hx}[f_{t}]+W_{hh}h_{t-1}+b_{h}),\label{equ:lstm_hidden}
\end{equation}
\begin{equation}
y_{t}=\sigma(W_{yh}h_{t}+b_{y}).\label{equ:lstm_output}
\end{equation}

\begin{figure*}
	\centering \includegraphics[width=12cm]{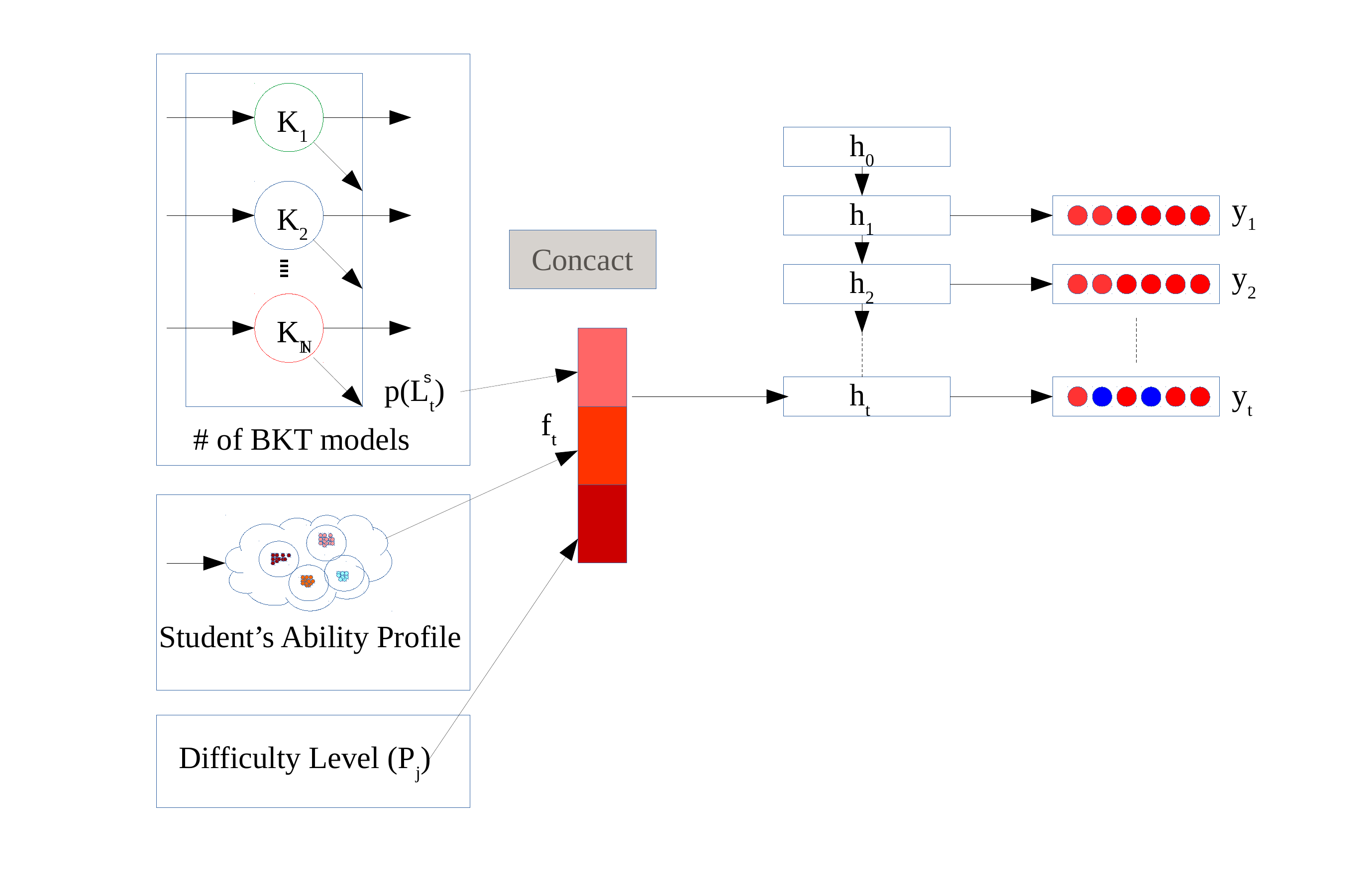} \caption{Architecture of BKT-LSTM.}
	\label{fig:BKT-LSTM} 
\end{figure*}

Both tanh and the sigmoid function are applied el$r_{t+1}$ement wise
and parameterized by an input weight matrix $W_{hx}$, recurrent weight
matrix $W_{hh}$, and output weight matrix
$W_{yh}$. Biases for latent and output units are represented by
$b_{h}$ and $b_{y}$. The input features~$f_{t}=\{P(s_t),ab_{z},PD(P_j)\}$ contains assessed skill mastery $M(s_t)$ of student $i$ on skill $s$ at time $t$, temporal ability profile $ab_{z}$ indicates the cluster ID~$Cluster(stu_{z}^{i})$, the temporal ability profile $ab_z$ of student $i$ at current time interval~${z}$, and $PD(P_j)$ represents problem difficulty of $P_j$ at time $t$. (for brevity we omit indexing all terms by student $i$ from here on). 
The output $y_t$ is represents the predicted probability that the student would answer that particular problem with associated skill correctly. Thus the prediction of  of problem associated with skill $s_{t}$ can be retrieved from vector of $y_t$ as described in Fig\ref{fig:BKT-LSTM}. 
\subsubsection{}{Optimization:}
To improve the predictive performance of RNN based models, we trained with the cross-entropy loss $l$ between probability of getting correctly of problem $j$ with skill $s$: $p_t = p(P_{j}^s) $ and actual response $r_t$ for all RNN based models as follows:

\begin{equation}
l=\sum_{t}(r_t \log p_t + (1-r_t) \log(1-p_t)),\label{equ:loss}
\end{equation}

It is to prevent overfitting during training, dropout was applied to $h_t$ when computing the prediction $y_t$, but not when computing the next hidden state $h_{t+1}$.

\section{Experimental Study}

\label{sec:experimental-study}

\subsection{Datasets}

\label{sec:dataset}

In order to validate the proposed model, we tested
it on three public datasets from two distinct tutoring scenarios in
which students interact with a computer-based learning system in the
educational settings: 1) ASSISTments \footnote{\url{https://sites.google.com/site/assistmentsdata/}}: 
 an online tutoring system that was first created in 2004 which
engages middle and high-school students with scaffolded hints in their
math problem. If students working on ASSISTments answer a problem
correctly, they are given a new problem. If they answer it incorrectly,
they are provided with a small tutoring session where they must answer
a few questions that break the problem down into steps. Datasets are as follows: ASSISTments 2009-2010 (skill builder), ASSISTments 2014-2015.
2) Cognitive Tutor. Algebra 2005-2006 \footnote{\url{https://pslcdatashop.web.cmu.edu/KDDCup/downloads.jsp}} :is a  development dataset released in KDD Cup 2010 competition from Carnegie Learning of PSLC DataShop.
\begin{table}
	\caption{Overview of datasets}
	
	\label{tab:data} 
	\begin{centering}
		\begin{tabular}{|c|r|r|r|r|}
			\hline 
			\multirow{3}{*}{Dataset} & \multicolumn{4}{c|}{Number of} \tabularnewline
			\cline{2-5} 
			& \multicolumn{1}{c|}{Skills}  & \multicolumn{1}{c|}{Problems} & \multicolumn{1}{c|}{Students}& \multicolumn{1}{c|}{Records}  \tabularnewline
			\hline 
		    Algebra & 437 & 15663 & 574  & 808,775   \tabularnewline
			\cline{1-1} 
			ASS-09 & 123 & 13002  & 4,163  & 278,607   \tabularnewline
			ASS-14 & 100 & NA & 19,840  & 683,801  \tabularnewline
			\hline 
		\end{tabular}
		\par\end{centering}
	
	\vspace{-2mm}
\end{table}
 For all datasets, only first correct attempts to original problems are considered in our experiment. We remove data with missing values for skills and problems with duplicate records. To the best of our knowledge, these are the largest publicly available knowledge tracing datasets.

In this experiment, we assumed every 20 attempts made by student as a time interval for that student. The total number of temporal values for student's learning ability used in our experiment is 8 (7 clusters and 1 for initial ability before evaluation in initial time interval for all students).  Five fold cross-validations are used to make predictions on all datasets. Each fold involves randomly splitting each dataset
into 80\% training students and 20\% test students of the each datasets. For Input of DKVMN, Initial values in both key and value memory are learned in training process. For other models, one hot encoded method is applied. Initial values in value memory represents for initial knowledge state as prior difficulty for each skill and fixed in testing process.

We compare next problem student performance prediction of our model with state-of-the-art models mentioned in above: Bayesian extensions of BIRT~\cite{wilson2016back}, BKT~\cite{corbett1994knowledge}, PFA~\cite{pavlik2009performance}, DKT~\cite{piech2015deep}, DKVMN~\cite{zhang2017dynamic}. But we do not compare with other variant models, because those are more or less similar and do not show significant difference in performance. We implement the all NN models with Tensorflow and LSTM based models share same structure of  fully-connected hidden nodes for LSTM hidden layer with the size of 200 for DKT. For speeding up the training process, mini-batch stochastic gradient descent is used to minimize the loss function. The batch size for our implementation is 32. We train the model with a learning rate 0.01 and dropout is also applied for avoiding over-fitting.We set the number of epochs into 100. All these models are trained and tested on the same sets of training and testing students. For BKT, we learn models for each skill and make predictions separately  and the results for each skill are averaged.

\subsection{Results}

\begin{table}
	\begin{center}
		\caption{AUC result for all tested datasets}\label{tab:exp1}
		\begin{tabular}{|cccc|}
			\hline
			
			\multirow{3}{*}{Models} & \multicolumn{3}{c|}{Datasets} \\
			\cline{2-4}
			&  ASS-09& ASS-14 & Algebra        \\
			\hline
			BIRT & 0.75 & 0.67 &  0.81   \\  
			
			PFA &  0.70 & 0.69 &  0.76 \\
			\hline
			BKT & 0.65  & 0.61  & 0.64 \\  
			
			DKT & 0.72  & 0.70 & 0.78   \\
			
			DKVMN &  0.71 & \textbf{0.71} &  0.78\\
			\hline
			
			\textbf{BKT-LSTM} & \textbf{0.80} & \textbf{0.71}  & \textbf{0.85} \\
			\hline

		\end{tabular}  
	\end{center}	
\end{table}

In Table \ref{tab:exp1}, BKT-LSTM performs significantly better than
state-of-the-art models in tested datasets. On the Algebra dataset,
compared with the DKT and DKVMN which has an maximum test AUC of 78, 0.81 in IRT, 0.76 in PFA and 0.64 for BKT, but BKT-LSTM has 0.85 with a notable gain of 10\% to original DKT and DKVMN .In ASSISTments09 dataset, BKT-LSTM also achieves about 10\% gain with maximum test AUC=0.80 when compared to DKT with AUC=0.72 and DKVMN has AUC=0.71. On the ASSISTments14 dataset, BKT-LSTM achieved maximum test AUC=0.70 only. In the latest ASSISTments14 dataset (which contains more students and less data compared to other three datasets and without problem information) and a bit lower to DKVMN.

\begin{table}
	\begin{center}
		\caption{RMSE result for all tested datasets}\label{tab:exp2}
		\begin{tabular}{|cccc|}
			\hline
			
			\multirow{3}{*}{Models} & \multicolumn{3}{c|}{Datasets} \\
			\cline{2-4}
			&  ASS-09 & ASS-14 & Algebra        \\
			\hline
			BIRT  & 0.44  & 0.44  & 0.37   \\  
			
			PFA & 0.45 & 0.42 & 0.39   \\
			\hline
			BKT &  0.47 & 0.51  & 0.44 \\  
			
			DKT & 0.45  & 0.42  & 0.38   \\
			
			DKVMN & 0.45 & \textbf{0.42} & 0.38 \\
			\hline
			
			\textbf{BKT-LSTM} & \textbf{0.41}  & \textbf{0.42}   & \textbf{0.35}  \\
			\hline

		\end{tabular}  
	\end{center}	
\end{table}

In Table \ref{tab:exp2}, when we compare the models in term of RMSE,
BKT is minimum 0.46 in ASSISTments09, and ASSISTments14, and 0.44 in Algebra. RMSE results in all dataset with maximum of 0.40 for BKT-LSTM while all other models are no more than 0.42. According to these results, BKT-LSTM  shows better performance than DKT, DKVMN and significantly than other models  in Algebra, ASSISTments09, ASSISTments14.

\subsection{Ablation studies}\label{sec:comp-across-diff}

The results have so far suggest there may be different impacts of each factor on the predictive performance of NN based models, and in particular the impact of item difficulty.  This question is further analyzed in this section. 

We compare our BKT-LSTM model through an ablation study with following factor combinations:
\begin{itemize}
	\item BKT-LSTM-1: skill mastery.
	\item BKT-LSTM-2: skill mastery and ability profile.
	\item BKT-LSTM-3: skill mastery and problem difficulty.
	\item BKT-LSTM-4: skill mastery, ability profile and problem difficulty.

\end{itemize}
In BKT-LSTM-1, it takes only skill mastery into account for student performance prediction which achieve higher performance than original BKT and have lower performance than DKT (where DKT takes only binary values of student previous interaction). When BKT-LSTM-2 takes skill mastery and ability profile of a student, it shows similar performance as DKT and much higher than BKT. It shows better performance than DKT while BKT-LSTM-3 takes skill mastery and problem difficulty. Our proposed model  BKT-LSTM-4 shows better performance than any state-of-the-art methods compared in this experiment. We conducts our analysis through Area Under Curve (AUC),  Root Mean Squared Error (RMSE)  and the Square of Pearson correlation ($r^2$).

\begin{table}
	\begin{center}
		\caption{AUC result for ablation study}\label{tab:exp3}
		\begin{tabular}{|cccc|}
			\hline
			
			\multirow{3}{*}{Models} & \multicolumn{3}{c|}{Datasets} \\
			\cline{2-4}
			&  ASS-09 & ASS-14 & Algebra        \\
			\hline
			
			BKT-LSTM-1 &0.686  & 0.680   & 0.730 \\  
			
			BKT-LSTM-2 & 0.720   & 0.701   & 0.743    \\
			
			BKT-LSTM-3 & 0.792   & 0.702  & 0.849 \\
			
			BKT-LSTM-4 &  0.802 &    0.707  & 0.851   \\
			\hline

		\end{tabular}  
	\end{center}	
\end{table}

\begin{table}
	\begin{center}
		\caption{RMSE result for ablation study}\label{tab:exp4}
		\begin{tabular}{|cccc|}
			\hline
			
			\multirow{3}{*}{Models} & \multicolumn{3}{c|}{Datasets} \\
			\cline{2-4}
			&  ASS-09 & ASS-14 & Algebra        \\
			\hline
		
			BKT-LSTM-1 &  0.451  & 0.426   & 0.398  \\  
			
			BKT-LSTM-2 &  0.439   & 0.421   & 0.394    \\
			
			BKT-LSTM-3 & 0.416 &  0.421 & 0.355  \\
			
			BKT-LSTM-4 & 0.409 &  0.420  & 0.353   \\
			\hline

		\end{tabular}  
	\end{center}	
\end{table}

\begin{table}
	\begin{center}
		\caption{$r^2$ result for ablation study}\label{tab:exp5}
		\begin{tabular}{|cccc|}
			\hline
			
			\multirow{3}{*}{Models} & \multicolumn{3}{c|}{Datasets} \\
			\cline{2-4}
			&  ASS-09 & ASS-14 & Algebra        \\
			\hline
			
			BKT-LSTM-1 & 0.095  &  0.079  & 0.146  \\  
			
			BKT-LSTM-2 & 0.142  & 0.100  &  0.165  \\
			
			BKT-LSTM-3 & 0.228 & 0.102  & 0.32  \\
			
			BKT-LSTM-4 & 0.255 &  0.106  & 0.328  \\
			\hline

		\end{tabular}  
	\end{center}	
\end{table}

The results are reported in tables \ref{tab:exp3}, \ref{tab:exp4} and \ref{tab:exp5}. Although ability profile provides indication in student ability evolution in student long-term learning process. It has mostly improved in student performance prediction around 2 to 4 \% in BKT-LSTM 2 and 4 compared to BKT-LSTM 1 and 3. Results from each model with different combination of factors show us which components provide more information in student performance prediction among various datasets. In there, we can also see that the problem difficulty factor is the most influential factor to student performance prediction. When we apply problem difficulty to BKT-LSTM, it increases around 8 to 10 \% respectively. It also reflects that is why BIRT shows better performance in student performance prediction among other models (except neural networks based models with problem difficulty see Tables \ref{tab:exp1}, \ref{tab:exp2},  \ref{tab:exp3} and \ref{tab:exp4}). For ASSISTments14 dataset, all of the models (including BIRT, see Tables \ref{tab:exp1} and \ref{tab:exp2}) show similar performance, because it does not have problem information in dataset and we cannot calculate problem difficulty for each problem.

\section{Contribution}\label{sec:conclusion}

Neither BKT nor DKT method is a
perfect method for assessing knowledge mastery (knowledge tracing) and student performance prediction at the same time, but, when combined, they account for
different aspects of a  model and provide us better understanding with these meaningful features. 
 We proposed a student model called BKT-LSTM with three meaningful components: student's skill mastery (probability of learning a skill),  ability profile (learning transfer of a student) and problem difficulty to predict student performance. Not like DKT and DKVMN, which only take student past interaction (binary values) and learn all information in hidden state. BKT-LSTM can predict the student performance by summarizing information from skills, problems and student ability profiles. Experiments with three datasets show that the proposed model performs statistically and significantly better in predictive performance than state-of-the-art KT models. Dynamic evaluation of student’s ability profile at each time interval and problem difficulty at each time step plays critical roles and helps BKT-LSTM to capture more variance in the data, leading to more accurate and personalized performance predictions.

\end{document}